# A CONFIGURABLE PYTHONIC DATA CENTER MODEL FOR SUSTAINABLE COOLING AND ML INTEGRATION

**Avisek Naug, Antonio Guillen, Ricardo Luna Gutierrez, Vineet Gundecha, Sahand Ghorbanpour, Sajad Mousavi, Ashwin Ramesh Babu, Soumyendu Sarkar**\*

Hewlett Packard Enterprise (Hewlett Packard Labs)

```
avisek.naug, antonio.guillen, rluna, vineet.gundecha, sahand.ghorbanpour,
sajad.mousavi, ashwin.ramesh-babu, soumyendu.sarkar @hpe.com
```

## ABSTRACT

There have been growing discussions on estimating and subsequently reducing the operational carbon footprint of enterprise data centers. The design and intelligent control for data centers have an important impact on data center carbon footprint. In this paper, we showcase PyDCM, a Python library that enables extremely fast prototyping of data center design and applies reinforcement learning-enabled control with the purpose of evaluating key sustainability metrics including carbon footprint, energy consumption, and observing temperature hotspots. We demonstrate these capabilities of PyDCM and compare them to existing works in EnergyPlus for modeling data centers. PyDCM can also be used as a standalone Gymnasium environment for demonstrating sustainability-focused data center control.

## 1 INTRODUCTION

Enterprise data centers (DC) generate a massive carbon footprint relative to commercial buildings of similar sizes (10). This can be attributed to the high density of servers and associated CPU workloads that are scheduled throughout the day. The heat generated by these servers also needs to be ducted through an elaborate Heating Ventilation and Air Conditioning (HVAC) system from the IT Room to the outside environment through computer room air conditioning (CRAC) units, chillers, evaporators, pumps, fans, and cooling towers. The HVAC system also consumes a large amount of energy and has a significant carbon footprint depending on the workload of the data center, the weather conditions, and the grid renewable and non-renewable energy mix termed as the grid carbon intensity (CI).

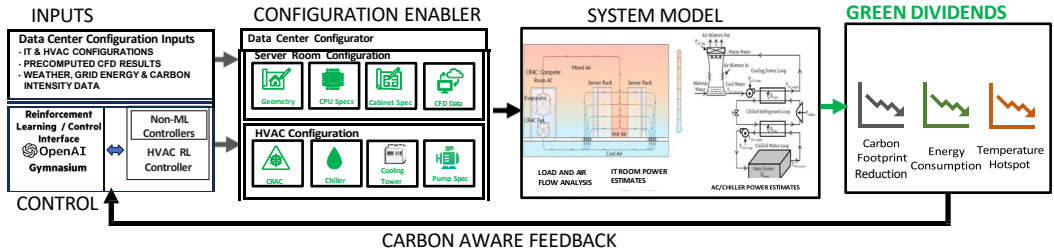

Figure 1: PyDCM Data Center Modeling Framework.

Software that enables designers to prototype thermal-efficient designs and test carbon-efficient control for data centers allows enterprises to meet their carbon neutrality target. Researchers have been relying on tools such as EnergyPlus (3) or Modelica (8) based models for design and control, which, in combination with interface software such as Sinergym (5) and PyFMU (1), allow for test of reinforcement learning (RL)-based controllers. However, there is considerable communication overhead between the modeling platforms and the RL controller, with the latter being commonly designed in Python programming language. Hence, researchers tend to develop such models in

---

\*Corresponding author

Python for ease of use. One such modeling approach, CityLearn (25), has been very popular among smart grid researchers.

Our work aims to provide the data center designer and machine learning communities with novel ways in Python to approach the sustainability target using both design and machine learning approaches. Specifically, the approach demonstrates the ability to generate custom data center configurations, study data center heat maps, and develop carbon-aware control applications using Deep reinforcement learning (DRL), that incentivize lower carbon footprint while optimizing the HVAC cooling setpoint for the IT room.

## 2  SYSTEM ARCHITECTURE

An overview of the proposed architecture with the proposed use cases is shown in figure 1. Inputs (*DC Configuration*) include parameters and settings for the various IT equipment and HVAC components, "supply" and "approach" (24) temperature results precalculated from a Computational Fluid Dynamics (CFD) (4) simulation, and local grid weather, energy, and carbon intensity (CI) data. All these parameters and configurations can be chosen by the framework users and automatically imported by the configuration reader.

Our proposed software follows an Object-Oriented Design (OOD) approach with the vectorized implementation of the thermal calculations (24) for the IT Models. The OOD approach allows the user to hierarchically design data center models that allow populating rooms with the required number and geometric configuration of IT cabinets and populate each cabinet with different types of servers/CPUs, each having their own processing power and fan power characteristics. The vectorized calculations scale with the number of servers/CPUs housed in the IT Cabinets, leading to faster simulation steps, which is essential for control approaches based on RL.

Next, we provide the detailed models implemented in the data center for the software:

**Data Center IT Model:** Let $\tilde{B}_t$ be the DC workload at time instant $t$. The spatial temperature gradient, $\Delta \mathbf{T}_{supply}$, given the DC configuration, is obtained from Computational Fluid Dynamics (CFD). For a given rack, the inlet temperature $T_{inlet,i}$ at $CPU_i$ is computed as:

$$T_{inlet,i,t} = \Delta \mathbf{T}_{supply,i} + T_{CRACsupply,t} \qquad (1)$$

where $T_{CRACsupply,t}$ is the CRAC unit supply air temperature. This value is chosen by an RL agent. Next, the CPU $j$ power curve $f_{cpu,j}(inlet\_temp, cpu\_load)$ and IT Fan power curve $f_{itfan,j}(inlet\_temp, cpu\_load)$ are implemented as linear equations based on (24). Given a server inlet temperature of $T_{inlet,i,t}$ at IT Cabinet $i$ and a workload of $\tilde{B}_t$ performed by all the $N_i$ CPUs in $i$, the total IT Cabinet $i$ power consumption ($P_{rack,i,t}$), and subsequently the total DC IT Power Consumption ($P_{datacenter,t}$) across all K cabiinets from $i = 1$ to $K$ can be calculated as follows:

$$P_{CPU,i,t} = \sum_{j=1}^{N_i} f_{cpu,j}(T_{inlet,i,t}, \tilde{B}_t) \qquad P_{IT\ Fan,i,t} = \sum_{j=1}^{N_i} f_{itfan,j}(T_{inlet,i,t}, \tilde{B}_t)$$

$$P_{rack,i,t} = P_{CPU,i,t} + P_{IT\ Fan,i,t} \qquad P_{datacenter,t} = \sum_i P_{rack,i,t}$$

The framework also provides an extensive set of models for the HVAC system (2) (*HVAC*) which can be subclassed for designing custom models by the user. The parameters may be set from existing parameter estimation methods to simulate models of actual HVAC components.

**HVAC Cooling Model:** Based on the DC IT Load $P_{datacenter,t}$, the IT fan airflow rate, $V_{sfan}$, air thermal capacity $C_{air}$, and air density, $\rho_{air}$, the rack outlet temperature $T_{outlet,i,t}$ for cabinet $i$ is estimated from (24) using:

$$T_{outlet,i,t} = T_{inlet,i,t} + \frac{P_{rack,k,t}}{C_{air} * \rho_{air} * V_{sfan}} \qquad (2)$$

In conjunction with the return temperature gradient information $\Delta \mathbf{T}_{return}$ estimated from CFDs, the final CRAC return temperature is obtained as:

$$T_{CRACreturn,t} = avg(\Delta \mathbf{T}_{return,i} + T_{outlet,i,t}) \tag{3}$$

We assume a fixed-speed CRAC Fan unit for circulating air through the IT Room. Hence, the total HVAC cooling load for a given CRAC setpoint $T_{CRACsupply,t}$, return temperature $T_{CRACreturn,t}$ and the mass flow rate $m_{crac,fan}$ is calculated as:

$$P_{cool,t} = m_{crac,fan} * C_{air} * (T_{CRACreturn,t} - T_{CRACsupply,t}) \tag{4}$$

To perform $P_{cool,t}$ the amount of cooling, the net chiller load for a chiller with Coefficient of Performance ($COP$) may be estimated as:

$$P_{chiller,t} = P_{cool,t} \left(1 + \frac{1}{COP}\right) \tag{5}$$

This cooling load is serviced by the cooling tower. Assuming a cooling tower delta as a function of ambient temperature $f_{ct\_delta}(T_{ambient,drybulb})$ (2), the required cooling tower air flow rate is calculated as:

$$V_{ct,air,t} = \frac{P_{chiller,t}}{C_{air} * \rho_{air} * f_{ct\_delta}(T_{ambient,drybulb})} \tag{6}$$

Finally, the Cooling Tower Load at a flow rate of $V_{ct,air,t}$ is calculated with respect to a reference air flow rate $V_{ct,air,REF}$ and power consumption $P_{ct,REF}$ from the configuration object:

$$P_{HVAC,cooling,t} = P_{ct,REF} * \left(\frac{V_{ct,air,t}}{V_{ct,air,REF}}\right)^3 \tag{7}$$

The goal of the DC HVAC RL agent is to minimize the total cooling energy and hence the carbon footprint by controlling the $A_{dc,t} = T_{CRACsupply,t}$ given the current CPU workload ($\hat{B}_t$), weather condition, grid CI ($CI_t$), UPS Battery SoC ($BatSoC$) and other related temporal and spatial information as outlined in the equations above.

## 3 SUSTAINABILITY METRICS IN PYDCM

PyDCM allows the user to track key performance indicators (KPI) related to sustainable data center operation.

**Energy Footprint** It depends on the energy demand of the data center which comprises the *IT Server Energy* $P_{it}(t, \theta_{it})$, *IT Fan Energy* $P_{fan}(t, \theta_{fan})$ and *HVAC Cooling Energy* $P_{cool}(t, \theta_{hvac})$. The $\theta_*$ parameters quantify the dependency of the design or control decisions for the data centers, while the temporal aspect is attributed to the incoming CPU workload and weather variables.

$$Energy\ Footprint(t, \theta_{it}, \theta_{fan}, \theta_{hvac}) = P_{it}(t, \theta_{it}) + P_{fan}(t, \theta_{fan}) + P_{cool}(t, \theta_{hvac})$$

**Carbon Footprint** It depends on the net energy demand of the data center *Energy Footprint* and the carbon intensity (CI) of the grid $CI(t)\ gCO_2/kwh$.

$$Carbon\ Footprint(t, \theta_{it}, \theta_{fan}, \theta_{hvac}) = CI(t) * Energy\ Footprint(t, \theta_{it}, \theta_{fan}, \theta_{hvac})$$

**Temperature Hotspot** It indicates the highest temperature sensed across the IT Rooms. It is affected by the geometric and hardware design choices of the data center. Efficient designs provide overall lower temperatures at the IT Cabinet server outlets.

$$T_{hotspot} = max(\mathbf{\Phi})$$

where $\mathbf{\Phi}$ is the $3D$ temperature distribution matrix of the data center for a given set of design, hardware, and HVAC setpoint ($\theta_{hvac}$) choices.

## 4 COMPARATIVE ANALYSIS OF PYDCM AND CURRENT ENERGY PLUS MODELS

### 4.1 COMPARISON WITH RL APPLICATIONS

We benchmarked PyDCM against the current data center implementations in EnergyPlus (24; 9) by focusing on three RL methods: `init`, `reset`, and `step`. The cumulative simulation times, combining `reset` and `step`, were also analyzed for different episode lengths: 7 and 30 days. The simulation time step was 15 minutes. All tests were carried out in a data center with two zones, as demonstrated in (24; 9). RL applications have been also successful in other domains (20; 21; 18; 16; 17; 11; 14; 12; 23; 19; 20; 23; 13; 15; 22).

**RL Method Time Analysis** In our evaluation, detailed in Table 1, PyDCM showcased significant improvement across the `init`, `reset`, and `step` RL methods. PyDCM's acceleration can be attributed to its utilization of vectorized and in-place computations for data center dynamics, which optimizes both memory and compute time.

**Total Simulation Time Analysis** The total simulation times for different episode lengths are summarized in Table 2. The individual improvements in the step and reset methods lead to cumulative improvements.

Table 1: Comparison of method timings between EnergyPlus and PyDCM. Mean ± std. dev. of 10 simulations.

| Method | EnergyPlus | PyDCM | Reduction (%) |
|---|---|---|---|
| `init`  | 1.05s ± 23.6ms   | 1.57ms ± 60.4μs  | 99.85 |
| `reset` | 2.67s ± 23.8ms   | 0.03ms ± 0.25μs  | 99.99 |
| `step`  | 0.46ms ± 98.38μs | 0.13ms ± 15.84μs | 71.33 |

Table 2: Total simulation time comparison for different RL episode lengths. Mean ± std. dev. of 10 simulations.

| Episode | EnergyPlus | PyDCM | Reduction (%) |
|---|---|---|---|
| 30 days | 3.33s ± 91.20ms | 0.34s ± 42.20ms | 89.79 |
| 7 days  | 2.64s ± 34.39ms | 0.09s ± 1.86ms  | 96.77 |

Table 3: Comparison of Performance Metrics for RL Environments. Mean ± std. dev. of 10 simulations.

| Metric | EnergyPlus | PyDCM | Reduction (%) |
|---|---|---|---|
| Wait. Time   | 1.48s ± 0.22s | 0.27s ± 0.48ms   | 81.55 |
| Sample Time  | 9.28s ± 0.51s | 3.95s ± 16.20ms  | 57.34 |

### 4.2 SCALABILITY

While we demonstrate the speedup compared to EnergyPlus on a limited configuration, to assess scalability, we conducted a series of simulations, progressively increasing the number of simulated CPUs and tracking the total simulation time. The derived results, illustrated in Figure 2, lead to two pivotal insights:

**Performance Enhancement with PyDCM:** PyDCM significantly outperforms the existing EnergyPlus implementation for data center simulations. Specifically, PyDCM can operate at speeds more than 40 times faster than EnergyPlus. When examining hyper-scale data centers—characterized by more than 10,000 CPUs (denoted with a vertical line as "Hyper-Scale DC" (6))—PyDCM is able to reduce the simulation times by a factor of 16.

**Underlying Assumptions in EnergyPlus Calculations:** An interesting pattern emerged when observing the consistent simulation time across different CPU counts in EnergyPlus. This behavior suggests that the EnergyPlus model might be built on certain assumptions. It could calculate the energy and thermal properties of a single CPU and then linearly scale (multiply) this base value by the total number of CPUs. While such a method can streamline calculations, it limits customization.

### 4.3 RESOURCE UTILIZATION ANALYSIS

In terms of system resources, while optimizing control using RLLib (7), EnergyPlus uses $18.20 GB$ of RAM, while PyDCM uses a slightly lower $16.84 GB$. Moreover, PyDCM's CPU utilization is more efficient, registering at $18.21\%$, as opposed to EnergyPlus's $20.64\%$. These experiments were conducted on a server equipped with a 48-core Intel Xeon 6248 CPU.

### 5.1 ENERGY FOOTPRINT
### 5 APPLICATIONS OF PYDCM

For a fixed data center design and server specifications, we trained an HVAC setpoint optimizer using a Deep Reinforcement Learning algorithm that minimizes the total energy consumption. When benchmarked against a standard ASHRAE Guideline 36 Controller (RBC) we obtain $7.36\%$ energy savings as shown in figure 3a.

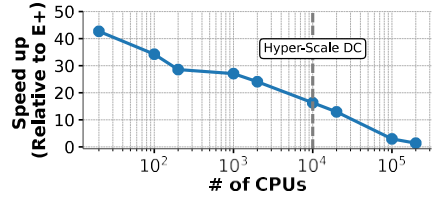

Figure 2: Simulation speed up relative to the current implementation of EnergyPlus (E+). * Hyper-scale data center consists of more than 10,000 CPUs. (6)

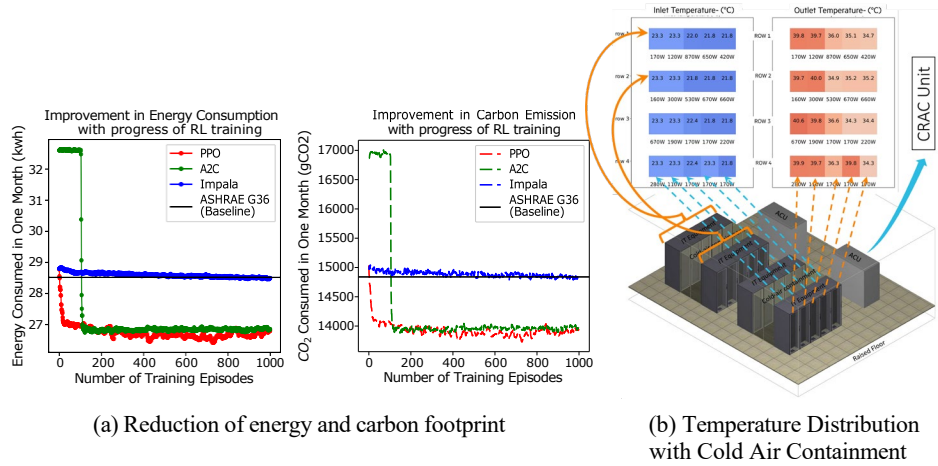

(a) Reduction of energy and carbon footprint

(b) Temperature Distribution with Cold Air Containment

Figure 3: Applications of PyDCM

## 5.2 CARBON FOOTPRINT REDUCTION

Similar to the energy reduction problem, when benchmarked against a standard ASHRAE Guideline 36 Controller (RBC) we obtain $7.23\%$ carbon footprint savings (figure 3a).

## 5.3 TEMPERATURE HOTSPOT ESTIMATION

Given a specific arrangement of the IT Cabinets and choices of servers, PyDCM helps evaluate the temperature distribution at the inlets and outlets of the cabinets. This is highlighted in figure 3b for a cold containment arrangement of a simple 2-row data center with 5 IT cabinets in each row.

## 6 CONCLUSION

In this paper, we developed a data center modeling and control-enabling framework. We demonstrated it resource effectiveness compared to current standards and its application in achieving sustainable data center operations.